\pdfoutput=1
\documentclass[runningheads,a4paper]{llncs}
\usepackage{cite}
\usepackage{amssymb}
\usepackage{graphicx}
\usepackage{multirow}
\usepackage{epstopdf}
\usepackage{url}
\usepackage{color}
\usepackage{booktabs}
\usepackage{amsmath}

\urldef{\mailsa}\path||
\urldef{\mailsb}\path||
\urldef{\mailsc}\path||
\newcommand{\keywords}[1]{\par\addvspace\baselineskip
\noindent\keywordname\enspace\ignorespaces#1}

\begin{document}

\mainmatter  % start of an individual contribution

\title{Attention-based Temporal Weighted Convolutional Neural Network for\\ Action Recognition}

\titlerunning{Attention-based Temporal Weighted CNN for Action Recognition}

\author{Jinliang Zang$^1$%
%\thanks{Please note that the LNCS Editorial assumes that all authors have used the western naming convention, with given names preceding surnames. This determines the structure of the names in the running heads and the author index.}%
\and Le Wang$^1$\and Ziyi Liu$^1$\and Qilin Zhang$^2$\and \\Zhenxing Niu$^3$\and Gang Hua$^4$\and Nanning Zheng$^1$}
\authorrunning{Lecture Notes in Computer Science: J Zang et al.}

\institute{$^1$Xi'an Jiaotong University, Xi'an, Shannxi 710049, P.R.China\\
$^2$HERE Technologies, Chicago, IL 60606, USA\\
$^3$Alibaba Group, Hangzhou, Zhejiang 311121, P.R.China\\
$^4$Microsoft Research, Redmond, WA 98052, USA
}

%\name{Xuhuan Duan$^{\star}$, Le Wang$^{\star}$, Changbo Zhai$^{\star}$, Gang Hua$^{\dagger}$, Qilin Zhang$^{\ddagger}$, and Nanning Zheng$^{\star}$}
%
%\address{$^{\star}$ Xi'an Jiaotong University, Xi'an, Shaanxi 710049, China \\
%	$^{\dagger}$ Microsoft Research, Redmond, WA 98052, USA \\
%	$^{\ddagger}$ HERE North America, Chicago, IL, 60606 USA}


\toctitle{Lecture Notes in Computer Science}
\tocauthor{Authors' Instructions}
\maketitle

\begin{abstract}
Research in human action recognition has accelerated significantly since the introduction of powerful machine
learning tools such as Convolutional Neural Networks (CNNs). However, effective and efficient methods for
incorporation of temporal information into CNNs are still being actively explored in the recent literature.
Motivated by the popular recurrent attention models in the research area of natural language processing,
we propose the Attention-based Temporal Weighted CNN (ATW), which embeds a visual attention model into a
temporal weighted multi-stream CNN. This attention model is simply implemented as temporal weighting yet
it effectively boosts the recognition performance of video representations. Besides, each stream in the
proposed ATW framework is capable of end-to-end training, with both network parameters and temporal weights
optimized by stochastic gradient descent (SGD) with backpropagation. Our experiments show that the proposed
attention mechanism contributes substantially to the performance gains with the more discriminative snippets
by focusing on more relevant video segments.
\end{abstract}

\section{Introduction}

Action recognition and activity understanding in videos are imperative elements of computer vision research. Over the last few years, deep learning techniques dramatically revolutionized research areas such as image classification, object segmentation\cite{wang2011automatic,wang2017video,long2015fully} and object detection \cite{ji20133d,karpathy2014large,zhang2012fast,s17061341,abeida2013iterative,carreira2017quo}. Likewise, Convolutional Neural Networks (CNNs) and Recurrent Neural Networks (RNNs) have been popular in the action recognition task \cite{simonyan2014two,yue2015beyond,wang2016temporal,carreira2017quo,donahue2015long,tran2015learning,cheron2015p,feichtenhofer2016convolutional,huang2018video}. However, various network architectures have been proposed with different strategies on the incorporation of video temporal information. However, despite all these variations, their performance improvements over the finetuned image classification network are still relatively small.

%\subsection{The challenges of action recognition}

Unlike image classification, the most distinctive property of video data is the variable-length. While Images can be readily resized to the same spatial resolution, it is difficult to subsample videos temporally. Therefore, it is difficult for the early 3D ConvNet \cite{ji20133d} to achieve action recognition performance on par with the sophisticated hand-crafted iDT \cite{wang2013action} representations.

In addition, some of the legacy action recognition datasets ({\em e.g.}, KTH \cite{schuldt2004recognizing}) only contain repetitive and transient actions, which are rarely seen in everyday life and therefore have limited practical applications. With more realistic actions included (with complex actions, background clutter and long temporal duration), the more recent action recognition dataset, {\em e.g.}, YouTube's sports, daily lives videos (UCF-101 \cite{soomro2012ucf101}) and isolated activities in movies (HMDB-51 \cite{kuehne2013hmdb51}), offer much more realistic challenges to evaluate modern action recognition algorithms. Therefore, all experimental results in this paper are based on the UCF-101 and HMDB-51 datasets.

\begin{figure}[t]
	\centering
	\includegraphics[width=0.7\textwidth]{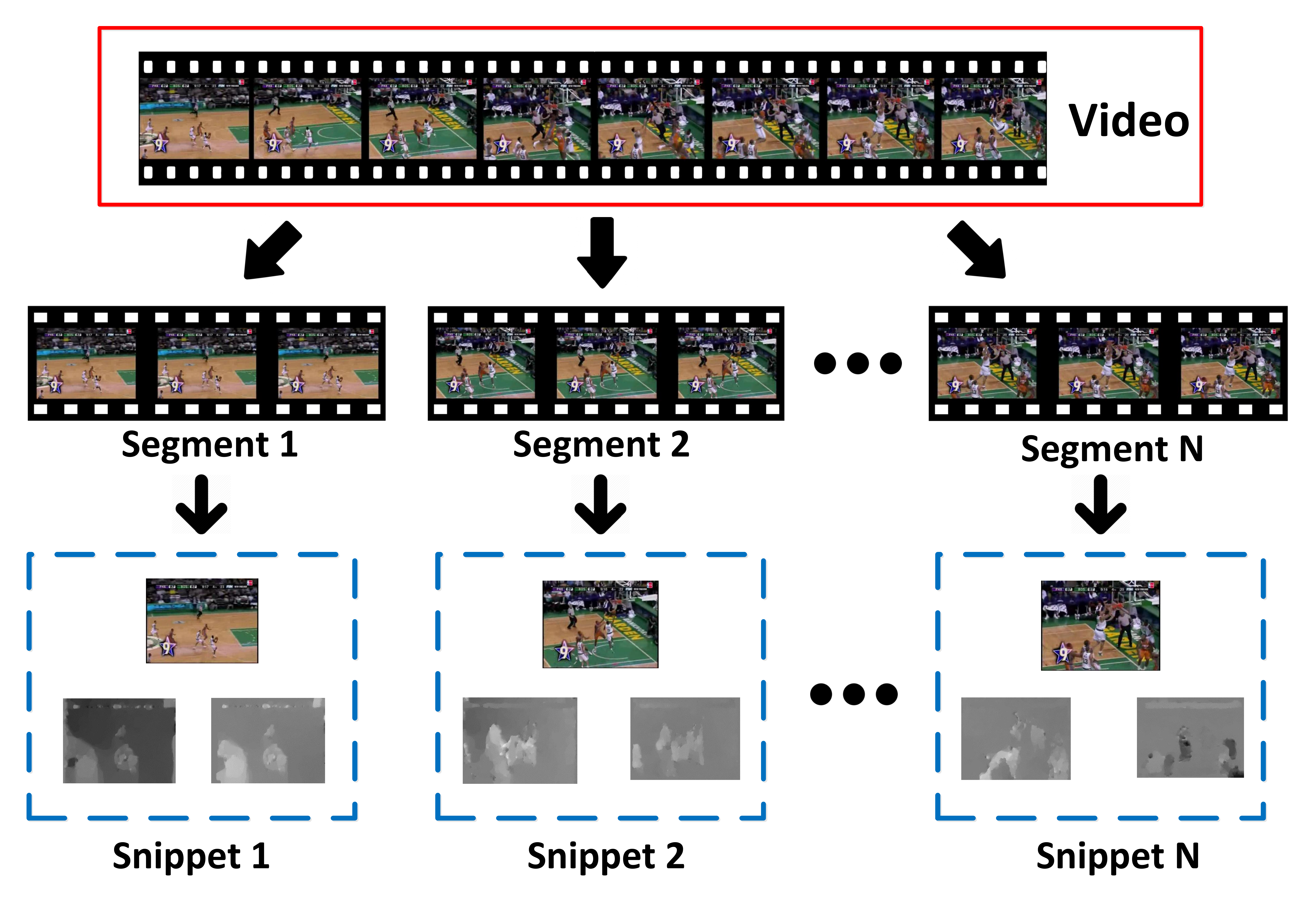}
	\caption{Snippet generation with a fixed target number ($N$) of chunks. A video is evenly portioned into $N$
	non-overlapping segments. Each segment contains approximately the same number of video frames.
	As shown above, $2$ additional modalities derived from RGB video frames are also included, {\em i.e.},
	optical flows and warped optical flows. RGB, optical flow and warped optical flow images sampled
	from the same segment are grouped in a snippet.}
	\label{fig:input}
\end{figure}

Previous multi-stream architecture, such as the two-stream CNN \cite{simonyan2014two}, suffers from a common drawback, their spatial CNN stream is solely based on a single image randomly selected from the entire video. For complicated activities and relatively long action videos (such as the ones in the UCF-101 and HMDB-51 datasets), viewpoint variations and background clutter could significantly complicate the representation of the video from a single randomly sampled video frame. A recent remedy was proposed in the Temporal Segment Network (TSN) \cite{wang2016temporal} with a fusion step which incorporates multiple snippets\footnote{Snippets are multi-modal data randomly sampled from non-overlapping video segments, see Fig.~\ref{fig:input}. Typically a video is divided into $1$ to $8$ segments. Segments are typically much longer than ``clips'' used by 3D CNN literature, {\em e.g.}, the $16$-frame clip in C3D \cite{tran2015learning}.}.

%\subsection{Our Model}
Inspired by the success of the attention model widely used in natural language processing \cite{luong2015effective} and image caption generation \cite{xu2015show,mnih2014recurrent}, the Attention-based Temporal Weighted CNN (ATW) is proposed in this paper, to further boost the performance of action recognition by the introduction of a benign competition mechanism between video snippets.  The attention mechanism is implemented via temporal weighting: instead of processing all sampled frames equally, the temporal weighting mechanism automatically focuses more heavily on the semantically critical segments, which could lead to reduced noise.  In addition, unlike prior P-CNN \cite{cheron2015p} which requires additional manual labeling of human pose, a soft attention model is incorporated into the proposed ATW, where such additional labeling is eliminated. Each stream of the proposed ATW CNN can be readily trained end-to-end with stochastic gradient descent (SGD) with backpropagation using only existing dataset labels.

%\subsection{Contribution}
The major contributions of this paper can be summarized as follows. (1) An effective long-range attention mechanism simply implemented by temporal weighting; (2) each stream of the proposed ATW network can be optimized end-to-end, without requiring additional labeling; (3) state-of-the-art recognition performance is achieved on two public datasets.

%\subsection{Paper Sections}

\section{Related Works}
Human action recognition has been studied for decades, which were traditionally based on hand-crafted features, such as dense trajectories \cite{wang2013action,wang2011action} and sparse space-time interest points \cite{laptev2005space}. In the past few years, CNN based techniques have revolutionized the image/video understanding \cite{ji20133d,simonyan2014two,karpathy2014large,yue2015beyond,wang2016temporal,carreira2017quo,donahue2015long,feichtenhofer2016convolutional,s17102421,s17061341}. Per the data types used for action recognition, deep neural networks based methods can be categorized into two groups: (1) RGBD camera based action recognition, usually with skeleton data and depth/3D point clouds information \cite{cheron2015p,wang2012mining,du2015hierarchical}; (2) conventional video camera based action recognition.

RGBD camera based action recognition offers 3D information, which is a valuable addition to the conventional RGB channels.
Such datasets are usually captured by the Microsoft Xbox One Kinect Cameras, such as The Kinetics dataset \cite{cheron2015p}.
Despite its obvious advantage, there are some limiting factors which restrict such model from wide applications. RGBD video
datasets are relatively new and labelled ones are not always readily available. A huge backlog of videos captured by
conventional RGB camcorders cannot be parsed by such methods due to modality mismatch \cite{zhang2015multi}. In addition, pure pose/skeleton based pipelines rarely achieve
recognition accuracy on par with RGB video frame based pipelines \cite{zhang2015can,zhang2015auxiliary}, making them more suitable for an auxiliary system to
existing ones.

Inspired by the success of computer vision with still RGB images, many researchers have proposed numerous methods for the
conventional RGB video camera based action recognition. Ji \textit{et al}. \cite{ji20133d} extend regular 2D CNN to 3D, with
promising performances achieved on small video datasets. Simonyan \textit{et al}. \cite{simonyan2014two} propose the two-stream
CNN, with each steam being a regular 2D CNN. The innovation is primarily in the second CNN steam, which parses a stack of optical
flow images that contain temporal information. Since then, optical flow is routinely used as the secondary modality in action
recognition. Meanwhile, 3D CNN has evolved, too. Tran \textit{et al}. \cite{tran2015learning} modified traditional 2D convolution
kernels and proposed the C3D network for spatiotemporal feature learning. Feichtenhofer \textit{et al}.
\cite{feichtenhofer2016convolutional} discovered one of the limiting factors in the two-stream CNN architecture, only a
single video frame is randomly selected from a video as the input of the RGB image stream. They proposed five variants of fusing
spatial CNN stream and two variants for the temporal steam. Additionally, Donahue \textit{et al}. \cite{donahue2015long}
developed a recurrent architecture (LRCN) to boost the temporal discretion. Consecutive video frames are loaded with redundant
information and noises, therefore, they argue that temporal discretion via LRCN is critical to action recognition. Some recent
literature also proposed new architectures with special considerations for temporal discretion
\cite{wang2016temporal,yao2015describing,gaidon2013temporal,huang2018video}.
\section{Formulation}%
Firstly, the temporally structured video representation is introduced, followed by the temporal attention model and the proposed ATW framework. 
\subsection{Temporally Structured Representation of Action}\label{sample}
{\em How do various CNN based architectures incorporate the capacity to extract semantic information in the time domain?} According to the previous two-stream CNN \cite{simonyan2014two} literature, there are generally $3$ sampling strategies: (1) dense sampling in time domain, the network inputs are consecutive video frames covering the entire video; (2) spare sampling one frame out of $\tau$ ($\tau \geq 2$) frames, i.e., frames at time instants $0,t,t+\tau,t+2\tau,\cdots,t+N\tau$ are sampled; (3) with a target number of $N$ segments\footnote{Typical $N$ values are from $1$ to $8$.}, non-overlapping segments are obtained by evenly partition the video into $N$ such chunks, as illustrated in Fig.~\ref{fig:input}.

As noted by \cite{feichtenhofer2016convolutional,donahue2015long,wang2016temporal}, the dense temporal sampling scheme is suboptimal, with consecutive video frames containing redundant and maybe irrelevant information, recognition performance is likely to be compromised.  For the sparse sampling strategy with $\tau$ intervals, the choice of $\tau$ is a non-trivial problem. With $\tau$ too small, it degrades to the dense sampling; with $\tau$ too large, some critical discriminative information might get lost. Therefore, the third sampling scheme with fixed target segments is arguably the advisable choice, given the segment number $N$ is reasonably chosen.

Suppose a video $V$ is equally partitioned into $N$ segments, {\em i.e.}, $V = \{S_k\}^N_{k=1}$, where $S_k$ is the $k$-th segment. Inspired by \cite{simonyan2014two,wang2016temporal,zhang2011fast}, multi-modality processing is beneficial. Therefore, three modalities (RGB video frame, optical flow image and warped optical flow image\footnote{As in \cite{ wang2013action}, warped optical flow is obtained by compensating camera motion by an estimated homography matrix.}) are included in our proposed ATW network.

One RGB video frame, five optical flow image and five warped optical flow images are randomly sampled from each segment $S_k$, as illustrated in Fig.~\ref{fig:input}, and respectively used as the inputs to the spatial RGB ResNet stream, temporal flow ResNet stream, and temporal warped flow ResNet stream, as shown in Fig.~\ref{fig:tan}. RGB, optical flow and warped optical flow images sampled from the same video segment are grouped in a snippet. Each snippet is processed by the proposed $3$-stream ATW network and a per-snippet action probability is obtained. After processing all snippets, a series of temporal weights are learned via the attention model, which are used to fuse per-snippet probabilities into video-level predictions.

\subsection{Temporal Attention Model}
\begin{figure}[t]
\centering
\includegraphics[width=1.0\textwidth]{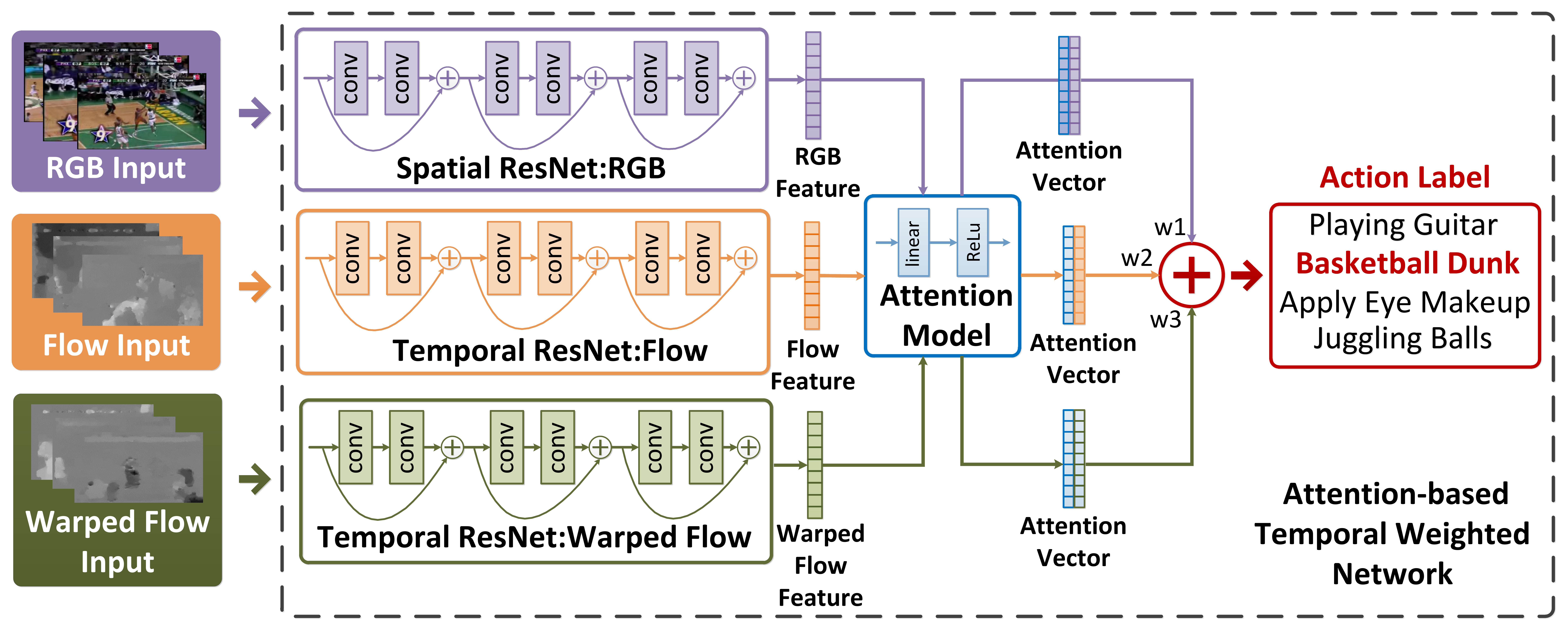}
\caption{Proposed ATW network architecture. Three CNN streams are used to process spatial RGB images,
temporal optical flow images, and temporal warped optical flow images, respectively. An attention model
is employed to assign temporal weights between snippets for each stream/modality. Weighted sum is used
to fuse predictions from the three streams/modalities.}
\label{fig:tan}
\end{figure}

The proposed ATW network architecture is presented in Fig.~\ref{fig:tan}. Our base CNN is the ResNet \cite{he2016deep}
or BN-Inception~\cite{ioffe2015batch}, which are both pretrained on the ImageNet dataset \cite{deng2009imagenet}. During
the training phase, every labeled input video $V$ is uniformly partitioned into $N$ segments, {\em i.e.},
$V=\{M_i^{RGB}, M_i^{F}, M_i^{WF}, y\}_{i=1}^N$, where $M_i^{RGB}, M_i^{F}, M_i^{WF}$ represent the RGB, optical flow and
warped optical flow images from the $i$th snippet, with $y$ being the corresponding action label. The $3$ CNN stream
($\mathcal{C}_{RGB}$, $\mathcal{C}_F$ and $\mathcal{C}_{WF}$) map each input to corresponding feature vector as
\begin{equation}
\begin{split}
&\mathcal{C}_{RGB}(M_i^{RGB}) = \mathbf{a}_i^{RGB}, \\
&\mathcal{C}_{F}(M_i^{F}) = \mathbf{a}_i^{F}, \\
&\mathcal{C}_{WF}(M_i^{WF}) =  \mathbf{a}_i^{WF}, \\
&i = 1, \cdots, N,
\end{split}
\end{equation}
where we call these $\mathbf{a}_{att}^{RGB}$, $\mathbf{a}_{att}^{F}$, $\mathbf{a}_{att}^{WF}$ action feature
vectors, and use $\mathbf{a}_i$ to represent any given one from the 3 modalities. Note that $w_i$ is the expected importance
value of the $i$th snippet relative to the entire video. Evidently, if $w_i \equiv \frac{1}{N}$, the attention model degrades
to naive averaging. The weight $w_i$ is computed by the attention model $f_{att}$ by a multi-layer perceptron conditioned on
the previous fully-connected hidden state ({\em i.e.}, $\mathbf{w}_{att}$). The value of weight $w_i$ decides which part of the
segments should to pay attention to. Formally, the attention model $f_{att}$ is defined as

\begin{equation}
e_i = f_{att}(\mathbf{w}_{att},\mathbf{a}_i) = \mathbf{w}_{att}^\mathbf{T}\mathbf{a}_i.
\end{equation}
The weight $w_i$ of each action vector is computed by
\begin{equation}
w_i =  \frac{\exp{e_i}}{\sum_j\exp{e_j}},
\end{equation}
where each $w_i$ are normalized by passing through a softmax function, which guarantees
they are positive with $\sum_i{w_i}=1$.
Finally, the attention mechanism $\varphi$ is implemented with a linear layer followed by a rectifier (ReLu), which serve as a
temporal weighting function that aggregates all the per-snippet prediction probabilities into a per-video prediction.
After training, the attention model obtains a set of non-negative weights $\{w_i\}^N_{i=1}$, so the weighted attention
feature is obtained by
\begin{equation}
\begin{split}
&\mathbf{A}_{att}^{RGB} = \varphi(\mathbf{a}_1^{RGB}, \cdots, \mathbf{a}_N^{RGB}) =
\sum_{i} w_i{\mathbf{a}_i^{RGB}}, \\
&\mathbf{A}_{att}^{F} = \varphi(\mathbf{a}_1^{F}, \cdots, \mathbf{a}_N^{F}) =
\sum_{i} w_i{\mathbf{a}_i^{F}}, \\
&\mathbf{A}_{att}^{WF} = \varphi(\mathbf{a}_1^{WF}, \cdots, \mathbf{a}_N^{WF}) =
\sum_{i} w_i{\mathbf{a}_i^{WF}}. \\
\end{split}
\end{equation}
For better readability, we give this new action feature vector $\mathbf{A}_{att}$ a name as
attention vector. To emphasize, the attention model directly computes a soft alignment, so that the gradient
of the loss function is trained by backpropagation.

\subsection{Implementation Details}

During the training phase, images from all three modalities (RGB, optical flow and warped optical flow) are cropped to $224\times224$. We employ cross modality pre-training \cite{wang2016temporal}. Firstly, the spatial stream (ResNet or BN-Inception) is pre-trained on the ImageNet image classification dataset. Subsequently, these pre-trained weights are used to initialize all $3$ streams in the ATW. Each stream of the proposed ATW is trained independently. We use a single frame ($1$) and a stack of ($5$) consecutive (warped) optical flow frame as inputs. Based on the standard cross-entropy loss function, the SGD algorithm is used with a mini-batch size of 128 videos. We use an initial learning rate of 0.001 for the spatial stream and 0.005 for both temporal streams. For spatial stream, the learning rate is multiplied by a factor of $0.1$ every 2000 iterations. For both temporal streams, the learning rate decay is divided into stages. Learning rates are multiplied by $0.1$ at iterations 12000 and 18000. All momentums are fixed at $0.9$.

During the testing phase, with each testing video, a fixed number of snippets (80 in our experiments) are uniformly sampled. We use weighted average fusion ($1, 1, 0.5$ for the spatial stream, optical flow stream, and warped optical flow stream, respectively) to generate a per-video prediction.

Pytorch \cite{paszke2017pytorch} is used in our experiments with optical flow and warped optical flow extracted via OpenCV with CUDA 8.0. To speed up training, 2 NVIDIA Titan Xp GPUs are used.

\section{Experiments}
%
%In this section, we evaluate the performance of the proposed ATW network in two public datasets against popular baselines.
%
%\subsection{Datasets and Baselines}
%
%\subsubsection{Trimmed Action Datasets.}
\noindent \textbf{Trimmed Action Datasets.} We evaluate our approach on two popular action recognition benchmarks, namely UCF-101
\cite{soomro2012ucf101} and HMDB-51 \cite{kuehne2013hmdb51}.
The UCF-101 dataset is one of the biggest action datasets containing 13320 videos
clips distributed in 101 classes. HMDB-51 dataset is a very challenging dataset with 6766 videos (3570 training and 1530 testing videos)
in 51 classes. Evaluation on these two trimmed datasets is performed using average accuracy over three training/testing splits.
%

%\subsubsection{Baselines.}
\noindent \textbf{Baselines.} Throughout the following section, we compare our proposed ATW network with the standard base architecture, mostly
two-stream with the single segment of a video ($N=1$). For network architecture, we choose the
traditional BN-Inception~\cite{ioffe2015batch} for comparison in experiments.
%

%\subsection{Attention Model}
%

%\subsubsection{Comparison with Different Consensus Functions.}
\noindent \textbf{Comparison with Different Consensus Functions.}
Firstly, we focus on comparing the attention model from two optional consensus
functions: (1) max segmental consensus; (2) average segmental consensus.
The max and weighted average consensus function is injected
at last fully-connected layer, whereas, average consensus can be used after softmax layer.
On the other hand, our attention model is set before softmax layer.
The experimental performance is summarized in Table~\ref{tb1:segmental}. We implement these
four segmental consensuses with the BN-Inception ConvNet~\cite{ioffe2015batch} on the first split of UCF-101.
The number of segmentation $N$ is set to $4$. We use the weighted average fusion of three-stream outputs to generate the video-level prediction.
Average segmental consensus performs slightly better than max function.
The best result is obtained by the proposed attention model. Thus it can be seen the usage of attention model
significantly improves temporal structure for action recognition.
\begin{table*}[t]
	\caption{Exploration of different segmental consensus functions on the UCF-101 dataset (split1).}
	\label{tb1:segmental}
	%\vskip 0.15in
	\begin{center}
		\begin{small}
			\begin{tabular}{lccr}
				\toprule
				Consensus Function~ & ~Spatial ConvNets~  & ~Temporal ConvNets~ & ~Two-Stream \\
				\midrule
				Max & $ 85.0\%  $& $ 86.0\% $ & $ 91.6\% $  \\
				Average &  $ 85.0\% $ &  $ 87.9\% $ & $ 93.4\% $ \\
				Attention Model & $\mathbf{86.7}\%$ & $\mathbf{88.3}\%$ & $\mathbf{94.6}\%$ \\
			 \bottomrule
			\end{tabular}
		\end{small}
	\end{center}
\end{table*}
%

%\subsubsection{Multi-Segment.}
\noindent \textbf{Multi-Segment.}
Specially, in Table~\ref{tb2:attention},
we use RGB modality for training on multi-segment temporal structure with BN-Inception ConvNet~\cite{ioffe2015batch}. Note that if $N<3$, the model is oversimplified, and the performance on UCF-101 (split1) has seriously degraded. The attention model boosts the mAP with $85.80\%$ on the UCF-101 dataset and $53.88\%$ on HMDB-51 dataset ($N$=4),
resulting from the successfully reduced training error. This comparison verifies the effectiveness of the soft attention mechanism on long-range temporal structure.
\begin{table*}[t]
	\caption{Experiments of different initialization strategies for initializing the attention
		layer's parameters and several traditional activation functions on the UCF-101 dataset (split1).
		Specifically, $weight=1/N(N=4)$ equivalent to average consensus.}
	\label{tb3:initialization}
	%\vskip 0.15in
	\begin{center}
		\begin{small}
			\begin{tabular}{lc|lc}
				\toprule
				Initialization & ~Spatial-Stream~ & Activation Function~ & ~Spatial-Stream \\
				\midrule
				$weight=1/N$~~ & $ 84.44\% $ & tanh & 84.91$ \% $ \\
				$weight=1$ &  $ 85.17\% $ & sigmoid & 85.29$ \% $ \\
				random gaussian~~~~~	 & $ \mathbf{85.80}\% $ & relu~~~~~~~ & $ \mathbf{85.80}\% $ \\
				\bottomrule
			\end{tabular}
		\end{small}
	\end{center}
\end{table*}
%

%\subsubsection{Parameters Initialization and Activation Function.}
\noindent \textbf{Parameters Initialization and Activation Function.}
As we train the proposed ATW CNN, an appropriate initialization of the attention
layer's parameters is crucial. We compare different initialization strategies: (1) the weight $w_i$ is set to 1,
bias is 0; (2) the weight $w_i$ is set to $\frac{1}{N}$, bias is 0; (3) random Gaussian distribution initialization.
In addition, on behalf of finding the most fitting activation functions,
we tested several traditional activation functions in the attention layer.
As shown in Table~\ref{tb3:initialization}, on the UCF-101 dataset, 1 for weight and 0 for bias initialization
achieves $85.80\%$ on the top of the three.
\begin{table*}[t]
	\caption{Exploration of ATW CNN with more number
		of segments on the UCF-101 dataset and HMDB-51 dataset (split1).}
	\label{tb2:attention}
	%\vskip 0.15in
	\begin{center}
		\begin{small}
			\begin{tabular}{lcccccccr}
				\toprule
				\multirow{2}{*}{Dataset~~~~~} & \multicolumn{8}{c}{Spatial-Stream Accuracy} \\
				\cline{2-9}
				& ~~N=1~~ & ~~N=2~~ & ~~N=3~~ & ~~N=4~~ & ~~N=5~~ & ~~N=6~~ & ~~N=7~~ & ~~N=8~~  \\
				\midrule
				UCF-101 & 83.33\% & 83.89\% & 84.80\% & $\mathbf{85.80\%}$ &  85.29\% & 85.21\% & 85.04\% & 85.55\% \\
				HMDB-51 & 50.07\% & 53.33\% & 53.01\% & $\mathbf{53.88\%}$  & 53.33\% & 55.36\% & 53.20\% & 53.14\% \\
				\bottomrule
			\end{tabular}
		\end{small}
	\end{center}
\end{table*}
%

%\subsection{Comparison with State-of-the-arts}

\noindent \textbf{Comparison with State-of-the-arts.} We present a comparison of the performance of Attention-based Temporal Weighted CNN and previous
state-of-the-art methods in Table~\ref{tbl:stoa}, on UCF-101 and HMDB-51 datasets. For that we used a
spatial ConvNet pre-trained on ImageNet, the temporal ConvNet was trained by cross-modality
pretraining. We choose ResNet~\cite{he2016deep} for network architecture.
As can be seen from Table \ref{tbl:stoa}, both our spatial and temporal nets
alone outperform the hand-crafted architectures of
~\cite{cai2014multi,wang2013action,peng2016bag,wang2016mofap}
by a large margin. The combination of
attention improves the results and is comparable
to the very recent state-of-the-art deep models
~\cite{simonyan2014two,wang2016temporal,sun2015human,tran2015learning,
wang2015action,varol2017long,zhu2016key,fernando2015modeling,ni2015motion}.

\begin{table*}[t]
 \caption{Comparison of our method with other state-of-the-art methods on the UCF-101 dataset and HMDB-51 dataset.}
 \label{tbl:stoa}
 \begin{center}
	 \begin{small}
	 \begin{tabular}{lc|lc}
	 \toprule
	 \multicolumn{2}{c|}{HMDB-51} & \multicolumn{2}{c}{UCF-101} \\
	 \midrule
	 \multicolumn{1}{l}{Model} & \multicolumn{1}{c|}{Accuracy} &
	 \multicolumn{1}{l}{Model} & \multicolumn{1}{c}{Accuracy} \\
		 \midrule
	 DT \cite{cai2014multi} & $ 55.9\% $ & DT \cite{cai2014multi} & $ 83.5\% $ \\
 	iDT \cite{wang2013action} & $ 57.2\% $ & iDT \cite{wang2013action} & $ 85.9\% $ \\
 	BoVW \cite{peng2016bag} & $ 61.1\% $ & BoVW \cite{peng2016bag} & $ 87.9\% $ \\
 	MoFAP \cite{wang2016mofap} & $ 61.7\% $ & MoFAP \cite{wang2016mofap} & $ 88.3\% $ \\
 	Two Stream \cite{simonyan2014two} & $ 59.4\% $ & Two Stream \cite{simonyan2014two} & $ 88.0\% $ \\
 	VideoDarwin \cite{fernando2015modeling} & $ 63.7\% $ & C3D \cite{tran2015learning} & $ 85.2\% $ \\
 	MPR \cite{ni2015motion} & $ 65.5\% $ & Two stream +LSTM \cite{yue2015beyond} & $ 88.6\% $ \\
 	$\mathrm{F_{ST}CN}$ (SCI fusion) \cite{sun2015human} & $ 59.1\% $ & $\mathrm{F_{ST}CN}$ (SCI fusion)
 	~\cite{sun2015human} & $ 88.1\% $ \\
 	TDD+FV \cite{wang2015action} & $ 63.2\% $ & TDD+FV \cite{wang2015action} & $ 90.3\% $ \\
 	LTC~\cite{varol2017long} & $ 64.8\% $ & LTC~\cite{varol2017long} & $ 91.7\% $ \\
 	KVMF~\cite{zhu2016key} & $ 63.3\% $ & KVMF~\cite{zhu2016key} & $ 93.1\% $ \\
 	TSN (3 modalities)~\cite{wang2016temporal}~~~~~~~~ & $ 69.4\% $ &
 	TSN (3 modalities)~\cite{wang2016temporal}~~~~~~~~ & $ 93.4\% $ \\
	\midrule
 	Proposed ATW & $ \mathbf{70.5}\% $ & Proposed ATW & $ \mathbf{94.6}\% $  \\
	 \bottomrule
   \end{tabular}
	 \end{small}
 \end{center}
\end{table*}
%

\begin{figure}[t]
	\centering
	\includegraphics[scale=0.23]{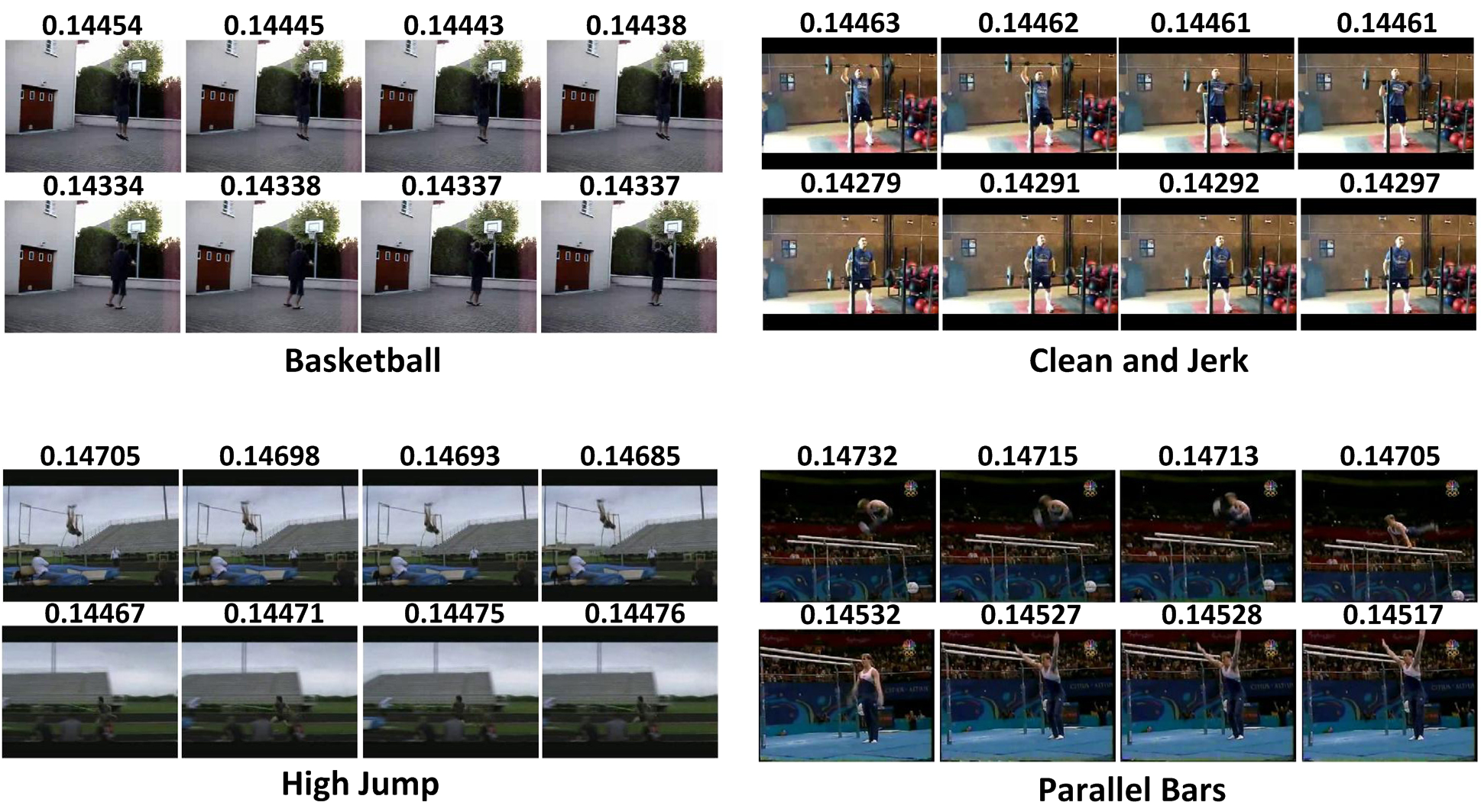}

	\caption{Visualization of the focus of attention on four videos from
		UCF-101 dataset over temporal dimension. The model learns to focus on the relevant parts.
		The attention weight is given on top of each image. The higher the attention weight ($w_i$) of the frame,
		the more critical to classify the action.}
	\label{fig:visualize}
	\vspace{-0.05in}
\end{figure}

%\subsection{Visualization}
\noindent \textbf{Visualization.} To analyze the ability of the proposed attention model and to select key snippets from long-range temporal multi-segment, we visualize what the proposed model has learned on frame-level, which can help to understand the operation of attention interpreting the feature activity. We test our model on several videos to acquire the expected value of the action feature, which can map this attention back to the input temporal dimension. Fig.~\ref{fig:visualize} presents what input images originally caused an attention value. The first row shows the top ranked four frames with their corresponding attention weights, and the second row shows the lowest ranked four frames. The attention model pays more attention to the relevant frames than irrelevant frames, and this means that the attention model always focuses on the foreground over time.

\section{Conclusion}

We presented the Attention-based Temporal Weighted Convolutional Neural Network (ATW), which is a deep multi-stream neural network that incorporates temporal attention model for action recognition. It fuses all inputs with a series of data-adaptive temporal weights, effectively reducing the side effect of redundant information/noises. Experimental results verified the advantage of the proposed method. Additionally, our ATW can be used for action classification from untrimmed videos, and we will test our proposed method on other action datasets in our future work.

%action datasets in our future work.

\section*{Acknowledgment}
This work was supported partly by NSFC Grants 61629301, 61773312, 91748208 and 61503296, China Postdoctoral Science Foundation Grant 2017T100752, and key project of Shaanxi province S2018-YF-ZDLGY-0031.

%\bibliographystyle{splncs}
%\bibliography{typeinst2}

%
%
%
\end{document}